\documentclass[11pt,a4paper]{article}

\usepackage{amssymb}
\usepackage{graphicx}
\usepackage{url}
\usepackage{amsmath}
\usepackage{algpseudocode}
\usepackage{algorithmicx}
\usepackage{color}

\usepackage{authblk}

\begin{document}

\title{
Reverse Engineering of Middleware for \\
Verification of Robot Control Architectures%\let\thefootnote\relax
\footnote{The final publication is available at Springer via \protect\url{http://dx.doi.org/10.1007/978-3-319-11900-7_27}}}

\author[1,2]{Ali Khalili\thanks{Ali.Khalili@edu.unige.it}}
\author[2]{Lorenzo Natale\thanks{Lorenzo.Natale@iit.it}}
\author[1]{Armando Tacchella\thanks{Armando.Tacchella@unige.it}}
\affil[1]{\scriptsize{DIBRIS, Universit\`a degli Studi di Genova,
Via Opera Pia 13 -- 16145 Genova -- Italy}}
\affil[2]{iCub Facility, Istituto Italiano di Tecnologia (IIT), 
Via Morego, 30 -- 16163 Genova -- Italy}

\renewcommand\Authands{ and }

\date{}
\maketitle

\begin{abstract}
We consider the problem of automating the verification of distributed
control software relying on publish-subscribe middleware. 
In this scenario, the main challenge is that software correctness
depends intrinsically on correct usage of middleware components, but
structured models of such components might not be available for
analysis, e.g., because they are too large and complex to  be described  
precisely in a cost-effective way. To overcome this problem, we
propose to identify abstract models of middleware as finite-state
automata, and then to perform verification on the combined middleware
and control software models. Both steps are carried out in a
computer-assisted way using state-of-the-art techniques in
automata-based identification and verification. Our main contribution
is to show that the combination of identification and verification is
feasible and useful when considering  typical issues that arise in the
implementation of distributed control software.  
\end{abstract}

\section{Introduction}
\label{sec:intro}
Publish-subscribe middleware such as ROS~\cite{quigley2009ros} and
YARP~\cite{yarpiCub2008} are becoming increasingly common in control
architectures of modern robots. The main advantage of using middleware
is that control modules can communicate seamlessly among each other
and with device-specific APIs, possibly across different computing
platforms. While operational scenarios for autonomous robots are
becoming increasingly complex --- see, e.g., the DARPA robotics
challenge~\cite{pratt2013darpa} --- the issue of dependability 
at all levels of a robot's architecture is getting more attention. 
In particular,  
if robots must be operated safely, control architectures must
be verified against various requirements, which include also software
specific properties, like deadlock or race avoidance. 
However, the task of verifying control software built on top
of some middleware cannot be accomplished unless a precise model of
the middleware is available, because a seemingly correct control code
can easily lead the robot to unwanted states if middleware
primitives are misused. An example of such case is when a sender
assumes buffered communication to a receiver, but the
channel is configured without buffering; if the sender expects
acknowledgment for every message, but some message is lost, then a
deadlock condition may ensue. 

Insofar a component of a control architecture is assigned precise
semantics, formal correctness verification is made possible, and many
control software fallacies can be spotted at design time.
However, developing a formal model can be difficult for
large and complex middleware like ROS or YARP.
A viable solution to this problem is to adopt
automata-based \textit{identification}  techniques -- see,
e.g.,~\cite{shahbaz2008} for a comprehensive list of references. The 
key idea is that the internal structure of a middleware
component can be inferred by analyzing its interactions with an
embedding context. Identification algorithms supply the component with
suitable input test patterns to populate a ``conjecture'' automaton by
observing the corresponding outputs; then, they check whether
the conjecture is behaviorally equivalent to the actual
component. When such an abstract model of the original component is
obtained, it can be used as a stub to verify software
components relying on it. This is where automata-based
\textit{verification} enters the scene. Given the inferred models of
middleware components, and a model of the control software relying on
them, Model Checking~\cite{que82a,cla86a} techniques provide an
automated way to check behavioral properties about the composition
of the models. In this way, confidence in a correct implementation of the
overall control architecture is increased, and problems can be spotted
before they cause expensive or even dangerous failures during robot's
operation.  

To demonstrate the effectiveness of our approach, we considered some
relatively simple, yet significant, examples of control code built on
top of YARP~\cite{yarpiCub2008}. Our choice is dictated by several
reasons, including a deep knowledge 
of the platform, and a fairly large installed base due to the adoption
of YARP as the standard middleware of  the humanoid
iCub~\cite{metta2010icub}. Moreover, YARP is a publish-subscribe
architecture quite similar to other middleware widely used in the
robotics community such as ROS. From the implementation point of view,
YARP is a set of libraries written in C++ consisting of more than 150K
lines of code. The purpose of YARP is to support modularity by
abstracting algorithms and the interface to the hardware and operating
systems. YARP abstractions are defined in terms of protocols. One of
the main features of YARP is to support inter-process
communication using a ``port'' abstraction. Our case studies focus
mostly on the identification of various concrete mechanisms underlying
this abstraction, e.g., buffered vs. non-buffered ports, and then to
check control code relying on such implementations.
Practical identification of different kinds of abstract models of
YARP ports is enabled by our tool AIDE (Automata IDentification
Engine)\footnote{AIDE, developed
  in C\#, is an open-source software: \url{http://aide.codeplex.com}}. 
 Model checking the composition of control
code and middleware is accomplished with the state-of-the-art tool
SPIN~\cite{holzmann2004spin}. The results obtained combining AIDE and SPIN,
albeit still preliminary, show that our approach is promising for the
identification and verification of control-intensive parts of the
code, i.e., those parts where the complexity of the code raises from
control flow rather than data manipulation.

The remainder of this paper is organized as follows. In
Section~\ref{sec:background}, a short summary of background and the
related works will be provided. Section~\ref{sec:casestudies}
introduces and motivates our YARP-based case
studies. Section~\ref{sec:exper} presents our experiments on
identification and verification. Finally, some concluding remarks and
possible directions of future works are given in Section~\ref{sec:conc}. 

%%%%%%%%%%%%%%%%%%%%%%%%%%%%%%%%%%%%%%%%%%%%%%%%%%%%%%%%%%%%%%%%%%

\section{Background}
\label{sec:background}

We define an \emph{interface automaton} (IA) as a quintuple 
$P = (I,O,Q,q_0,\rightarrow)$ where $I$ is a set of \emph{input actions},
$O$ is a set of \emph{output actions}, $Q$ is a set of states, $q_0\in Q$ is
the \emph{initial state} of the system, $\rightarrow\subset Q\times
(I\cup O) \times Q$ is the  \emph{transition relation}, and the sets
$O$, $I$ and $Q$ are finite, non-empty and mutually
disjoint. Our definition of IA is the same given
in~\cite{learningIOA2010}, which does not take into account
the possibility of formalizing hidden actions. Since we wish to infer 
IAs as models of middleware components, we can neglect such actions
without losing generality in our context. 
The set of all actions $A=I \cup O$ is the \emph{action signature} of
the automaton. Given a state $q \in Q$ and an action $a \in A$, we
define the \emph{next state function} $\delta:Q\rightarrow 2^{Q}$ as
$\delta(q,a) = \{ q' |  q\xrightarrow{a}q' \}$, where we write
$q\xrightarrow{a}q'$ to denote that $(q,a,q') \in \rightarrow$. An
action $a \in A$ is \emph{enabled} in a state $q \in Q$ if there
exists some $q'\in Q$ such that $q\xrightarrow{a}q'$, i.e.,
$|\delta(q,a)|\geq 1$. A state $q \in Q$ wherein all inputs are
enabled is \emph{input-enabled}, and so is an automaton wherein all
states are input-enabled. An input-enabled IA is also known as
\emph{I/O automaton}~\cite{ioAutomata1987}. 
Given a state $q \in Q$, the set $out(q) \subseteq Q$
of \emph{observable actions} is the set of all actions $a\in O$ where
$a$ is enabled in $q$. If $out(q) = \emptyset$, then  $q$ is
called \emph{suspended} or \emph{quiescent}. According
to~\cite{interfaceAutomata2001}, an \emph{execution fragment} of the
automaton is a finite alternating sequence of states and actions 
$u_0,a_0,u_1, \ldots, u_n$ such that $u_i \in Q$, $a_i \in A$ and
$u_i\xrightarrow{a_i}u_{i+1}$ for all $0 \leq i < n$. 
%%%%%%%%%%%%%%%%%%%%%%%%%%%%%%%%%%%%%%%%%%%%%%%%%%%%%%%%%%%%%%%%%%
\paragraph{Automata-based Inference}
Automata-based identification (also, automata learning) can be divided
into two wide categories, i.e., passive and active learning. In
\emph{passive learning}, there is no control over the observations received to
learn the model. In \emph{active learning}, the target system can be 
experimented with, and experimental results are collected to learn a
model. Whenever applicable, active learning is to be preferred because
it is computationally more efficient than passive learning --
see~\cite{kearns1994} for details. Furthermore, active learning is not
affected by a potential lack of relevant observations because it can
always query for them. However, active learning requires that the
target system is available for controlled experimentation, i.e., it
cannot be performed while the target is executing. 
The basic abstraction in active learning as introduced by Angluin
in~\cite{angluin1987}, is the concept of \textit{Minimally Adequate
Teacher} (MAT). In our case, it is assumed that a MAT exists and it can answer
two types of questions, namely \emph{output queries} and
\emph{equivalence queries}. An output query amounts to ask the MAT
about the output over a given input string, whereas equivalence
queries amount to compare a \emph{conjecture} about the abstract model
of a system with the system itself. The result of equivalence
queries is positive if the model and the system are behaviorally equivalent,
and it is a \emph{counterexample} in the symmetric difference of the
relations computed by the two automata, otherwise. In practice, 
since the system is unknown, equivalence queries must be approximated
by, e.g., model-based testing. Our tool AIDE
is a collection of learning algorithms for several abstract models,
including IAs. In particular, we use the Mealy machine inference
algorithm $L^+_M$~\cite{shahbaz2008} together with the approach
presented in~\cite{learningIOA2010} to identify IAs.
This model of identification is particularly suited in contexts where
the behavior of the system is jointly determined by its internal
structure, and by the inputs received from the environment -- also called
\emph{tester}. 
%%%%%%%%%%%%%%%%%%%%%%%%%%%%%%%%%%%%%%%%%%%%%%%%%%%%%%%%%%%%%%%%%%
\paragraph{Formal Verification}
Automata-based verification --- see, e.g.,~\cite{baier2008principles}
--- encompasses a broad set of algorithms and related tools, whose purpose
is to verify behavioral properties of systems represented as
automata. In particular, we consider algorithms and tools for Model
Checking~\cite{que82a,cla86a}. The basic idea behind 
automata-based verification technique is to exhaustively and
automatically  check whether a given system model meets a given
specification. In this approach, a property is specified usually in
terms of some temporal logic, and the system is given as some kind of
automaton.
In this work, we use SPIN~\cite{holzmann2004spin}, a generic verification
system that supports design and verification of asynchronous
process systems. In SPIN, the models are specified in a language  
called PROMELA (PROcess MEta LAnguage), and correctness 
claims can be specified in the syntax of standard linear temporal
logic (LTL). Several optimization techniques, including partial order 
reduction, state compression and bit-state hashing are developed to
improve performance of verification in SPIN. Details on the encoding
of IA into SPIN are given in Section~\ref{sec:exper}. Here, we give a
short overview on how verification works in SPIN and similar tools.
In SPIN, the global behavior of a concurrent system is obtained by
computing an asynchronous interleaving product of automata, where each
automaton corresponds to a single process. This means that, in
principle, SPIN considers every possible interleaving of the atomic
actions which every process is composed of. Technically, such product
is often referred to as the state space or \textit{reachability graph}
of the system. To perform verification, SPIN considers claims 
specified as temporal formulas. Typical claims include, e.g., safety
claims like ``some property is always/never true'', or liveness claims
like ``every request will be acknowledged''. Claims are converted into
B{\"u}chi automata, a kind of finite state automata whose acceptance
condition is suitable also for infinite words. The (synchronous)
product of automata claims and the automaton representing the 
global state space is again a B{\"u}chi automaton. If the
language accepted by this automaton is empty, this means that the
original claim is not satisfied for the given system. In other case,
it contains precisely those behaviors which satisfy the original
formula. Actually, the reachability graph is not computed up front
because, if $n$ is the number of state variables, the computation
would be exponential in $n$.
Rather, the composition of the two automata
is performed ``on the fly'', starting from the initial set of states
of the system, and then considering the reachable ones given the
process descriptions and the potential interleaving. 
SPIN terminates either by proving that some (undesirable) behavior is
impossible or by providing a counterexample match. 
%%%%%%%%%%%%%%%%%%%%%%%%%%%%%%%%%%%%%%%%%%%%%%%%%%%%%%%%%%%%%%%%%%
\section{Case Studies}
\label{sec:casestudies}

\begin{figure}[t!]
\scriptsize
\begin{center}
\begin{tabular}{ll}
\begin{minipage}{2.5in}
  \begin{algorithmic}[1]
  \State{Initialize buffered ports $Q_1$ and $Q_2$}
  \State{Connect $Q_1$ to $Q'_1$}
  \While{true}
  		\For{$i=1$ to $N$}  			
  			\State{Write message $m$ to $Q_1$}
  		\EndFor
  		\State{Read message from $Q_2$}
  \EndWhile  
  \end{algorithmic}  
\end{minipage}
& 
\begin{minipage}{2.5in}
  \begin{algorithmic}[1]
  \State{Initialize buffered ports $Q'_1$ and $Q'_2$}
  \State{Set the reading mode of $Q'_1$ as \textit{strict}}
  \State{Connect $Q'_2$ to $Q_2$}
  \While{true}
  		\For{$i=1$ to $N$}  			
  			\State{Read message $m$ from $Q'_1$}
  		\EndFor
  		\State{Create a message and write it to $Q'_2$}
  \EndWhile  
  \end{algorithmic} 
\end{minipage}
  \\ 
\end{tabular}
\end{center}
\caption{\scriptsize \label{fig:codeYARP}Case study 1: An example code \textit{Planner} (left) and  \textit{Controller} (right). 
	}
\end{figure}

Our motivation for this work is to to enable verification techniques
for robot control software which uses middleware modules. In this
section, we introduce two case studies. We focus on variations of the
well-known producer-consumer paradigm. The reason is that
similar situations are commonly found in robotic applications where 
loosely coupled modules are interconnected through publish-subscribe middleware and run concurrently.

\paragraph{Case Study 1.} We consider two
software components that exchange messages with loose
synchronization. A practical example is a \textit{Planner} ($P1$) that
generates a set of $N$ via points for a \textit{Controller} ($P2$). The
latter takes responsibility to execute each requests, in a variable
amount of time. The \textit{Planner} does not wait for execution of
the individual commands but rather sends all $N$ messages to
the \textit{Controller} and then waits for a synchronization packet
that signals the termination of the whole sequence. In a
publish-subscribe architecture this can be achieved using two
channels. The \textit{Planner} uses the first channel (between $Q_1$
and $Q'_1$) to send via points to the \textit{Controller}, then waits
for a message that acknowledges execution of the sequence from
the \textit{Controller} through the second channel (between $Q_2$ and
$Q'_2$). Since there is no synchronization, the buffering policy in the
connections can affect the correct behavior of the system. In this
application, the programmer assumes that connections are configured to
queue at least $N$ messages. This may or may be not true in YARP
where, by default, connections are configured to \emph{drop}
messages to reduce communication latencies\footnote{This policy may
seem counter-intuitive but it is fundamental for closed-loop
control}. The programmer therefore must override the default
configuration of the connections to ensure that messages are queued
and never dropped. Pseudo code for this scenario is given in
Figure~\ref{fig:codeYARP}. Given this code, we can see that, if
messages are dropped in the connections, a deadlock occurs. In
practice, the robot may not only fail to follow the desired
trajectory, but due to interpolation in the \textit{Controller}, it
may even end up in unsafe configurations.  

\begin{figure}[t!]
\scriptsize
\begin{center}
\begin{tabular}{ll}
\begin{minipage}{2.4in}
  \begin{algorithmic}[1]
  \State{\textbf{P1}}  
  \State{Initialize buffered ports $Q_1$}
  \State{Connect $Q_1$ to $Q'_1$}
  \For{$i=1$ to $N_1$}  			
		\State{//Do the job}
		\State{Send message $m$ to $Q_1$}
   \EndFor
  \end{algorithmic}  
  \begin{algorithmic}[1]
  \State{\textbf{P2}}  
  \State{Initialize buffered ports $Q_2$}
  \State{Connect $Q_2$ to $Q'_2$}
  \For{$i=1$ to $N_2$}  			
		\State{//Do the job}
		\State{Send message $m$ to $Q_2$}
   \EndFor
  \end{algorithmic}  
\end{minipage}
& 
\begin{minipage}{3in}
  \begin{algorithmic}[1]
  \State{\textbf{P3}}  
  \State{Initialize buffered ports $Q'_1$ and $Q'_2$}
  \State{Set the reading mode of $Q'_1$ and $Q'_2$ as strict}
  \For{$i=1$ to $N_3$}  			
		\State{Read message $m_1$ to $Q'_1$}
		\State{Read message $m_2$ to $Q'_2$}
		\State{//Do the job}
   \EndFor
  \end{algorithmic} 
\end{minipage}
  \\ 
\end{tabular}
\end{center}
\caption{\scriptsize \label{fig:codeYARP2}Case study 2: An example of
two producers ($P_1$ and $P_2$) and one consumer ($P_3$) which are
using YARP buffered port for their communication.} 
\end{figure}

\paragraph{Case Study 2.} In publish-subscribe architectures, sensory 
information and commands travel on distinct channels. It is therefore
common for components to receive information from multiple sources and
synchronize their activities on data received from such connections.
In this scenario, the consumer receives data from two
producers. A practical example is a grasping application. Here
the \textit{Tracker} ($P1$) identifies the 3D position of the object
in the work space (for example using stereo vision in the form of
$(x,y,z)$). This information reaches the \textit{Controller} ($P2$)
through the connection between $Q_1$ and $Q'_1$ which in turn
computes the torque commands to the motors. Another component ($P3$)
reads sensory data from a Force/Torque sensor placed in the kinematic
chain, and publishes it on a separate channel. The \textit{Controller}
relies on this information to detect collisions and control the force
exerted at the end-effector (connection between $Q_2$ and $Q'_2$). The 
programmer of the \textit{Controller} must decide how to read data
from both connections. The crucial point is that these connections can
become inactive. 
These might happen when no valid target is detected by
the \textit{Tracker}, or in situation where the \textit{Tracker} was
closed by the user or died unexpectedly.
By default, YARP defines that readers wait for data on a port (blocking
behavior). This allows tight synchronization and reduces latencies.
An inexperienced programmer may read data from both channels using the
default mode introducing an unexpected deadlock
when \textit{Tracker} does not produce data. The pseudo-code of this
scenario is presented in Figure \ref{fig:codeYARP2}. The connection
between $P1$ and $P2$ to $P3$ is implemented using buffered ports with
strict mode. To simulate the behavior where one of the producers stops
sending data we added a counter $N_i$ to the main loop of each
process. We verify the effects of different relative values of
$N_i$ on the overall behavior.

\begin{figure}[t!]
\begin{center}
\includegraphics[width=0.40\linewidth]{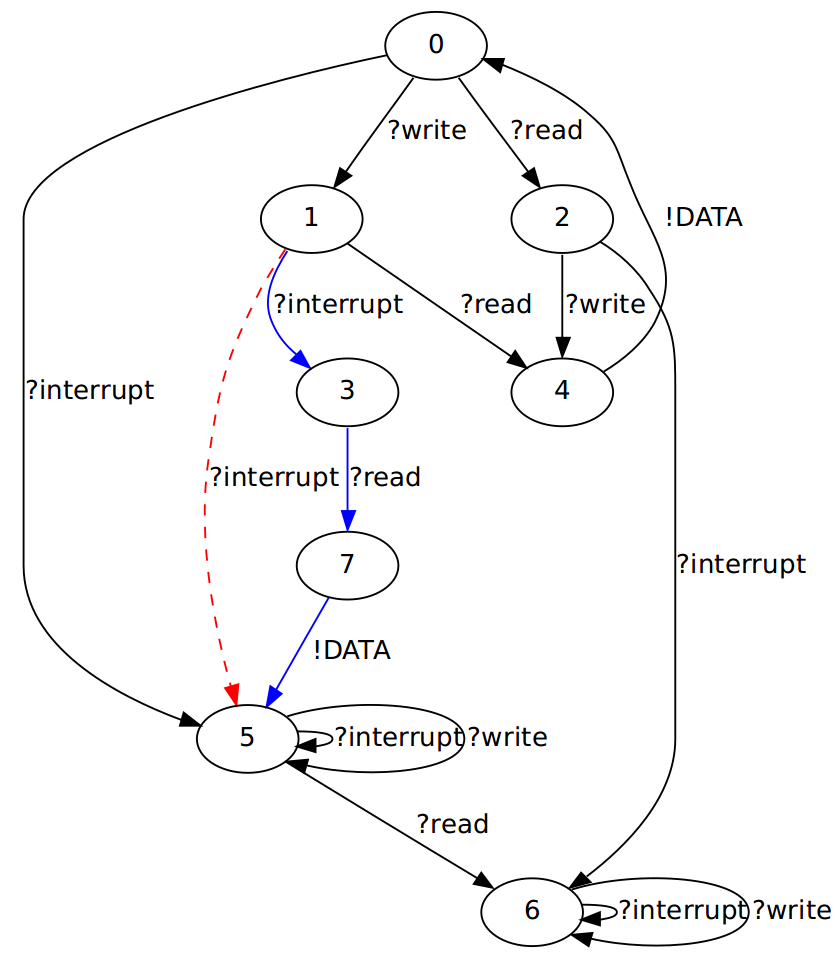}
\caption{\label{fig:interrupt}The identified model of a port with one
reader, one writer, and one thread which interrupts the write port. 
The expected model would feature the dotted transition --- from state $1$
into state $5$ --- but the actual model identified by AIDE and
implemented in YARP has the solid ones instead. Transitions labeled
with ``?'' and ``!'' represent input and output actions, respectively.} 
\end{center}
\end{figure}

In both the case studies describe above, we consider a combination of
identification and verification to be detailed in
Section~\ref{sec:exper}. However, we would like to point out that
identification alone is often useful to strengthen middleware, by
helping the discovery of corner bugs that are elusive in common usage
patterns. Out of many identification experiments that we conducted
with YARP in our preliminary work, we show in
Figure~\ref{fig:interrupt} the result of one which turned out as
a report in the YARP bug-tracking system. In this example, we consider
applying an interrupt method on a port. Interrupting a port is
supposed to unblock any blocked thread waiting for the port. The model in
Figure~\ref{fig:interrupt} is the one identified by AIDE for one port reader,
one port writer, and a thread which interrupts the writing port.
The model shows that interrupting a write port has been implemented so
that it unblocks future writes, but it waits for completion of the
current one, which was not the expected behavior. Indeed, this
specific behavior was not documented, and never occurred in YARP
practical applications, so it went unnoticed so far.

%%%%%%%%%%%%%%%%%%%%%%%%%%%%%%%%%%%%%%%%%%%%%%%%%%%%%%%%%%%%%%%%%%
\section{Experiments}
\label{sec:exper}

\begin{figure}[t!]
\begin{center}
\includegraphics[width=0.8\linewidth]{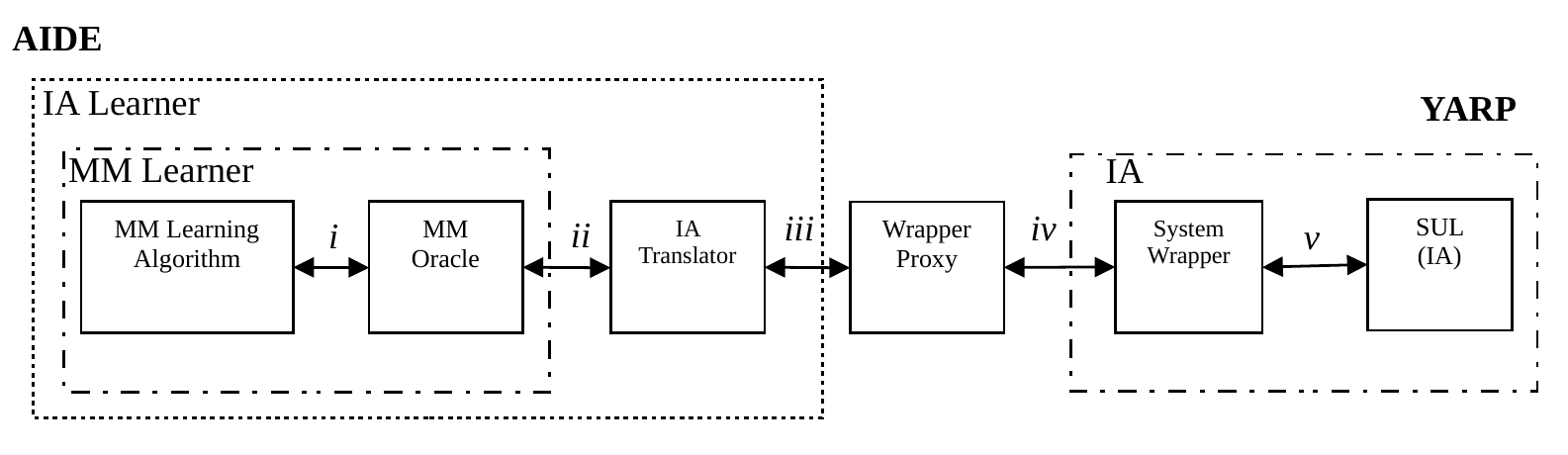}
\caption{\label{fig:config} Components of the
  learning procedure. The connections are $(i)$ queries asked to the MAT
  (MM Oracle), $(ii)$ bi-directional translation of interface automata and
  Mealy machines, $(iii)$ actions and events of the system, $(iv)$ the
  TCP connection to remote system-wrapper, $(v)$ the
  abstraction/concretization made by the system wrapper.} 
\end{center}
\end{figure}

Considering YARP port components as the system under learning (SUL)
to be modeled as IA, we use AIDE to identify abstract
models with different parameters. The configuration of components in
the inference procedure is presented in Figure~\ref{fig:config}. 
As we mentioned in Section~\ref{sec:background}, the basic inference
algorithm is $L^+_M$~\cite{shahbaz2008}, and it is implemented in the
module ``MM Learning Algorithm'' where ``MM'' stands for ``Mealy
Machine''. The algorithm relies on a software component, called ``MM
Oracle'' in Figure~\ref{fig:config}, whose task is to approximate the
behavior of a MAT on a real system. Together, these two modules are
the core of a Mealy machine inference program --- ``MM Learner'' in
Figure~\ref{fig:config}. Since we wish to identify YARP models as IAs, 
we connect a further component --- ``IA Translator'' in
Figure~\ref{fig:config} --- which implements  the approach presented
in~\cite{learningIOA2010} to identify IAs on top of a Mealy machine
learning algorithm. All these modules are part of AIDE, and they
collectively perform the task of ``IA learner''. The ``System
Wrapper'' component (in C++) bridges between the abstract alphabet on
the learner side, and the concrete alphabet of the SUL. It manages
different threads, handles the method calls in each thread, and the
abstraction of  messages ---  ``\textit{bottles}'' in YARP
terminology. To connect the wrapper to AIDE, 
we built a ``Wrapper Proxy'' which uses TCP/IP connections to
facilitate identifying systems remotely, possibly across different 
computing architectures.

\subsection{Identification of Ports in YARP}
\label{subsec:example}

\paragraph{Configuration of Ports} Considering a port connection,
we examine different parameters which affect its behavior. In
the case of a (standard) YARP port in a scenario with one sender and
one receiver, the type of communication is of a ``send/reply'' type,
wherein the sender and the receiver are tightly coupled. In the case
of buffered ports, the sender and the receiver enjoy more decoupling,
in the sense that YARP takes care of the lifetime of the objects being
transmitted through the port and it makes a pool of them, growing
upon need. By default, a buffered port keeps the most recent 
message only. Therefore, messages that come in between two
successive calls to read, might be dropped. If the so called
``strict'' mode is enabled, YARP will keep all received messages ---
like a FIFO buffer. Notice that, in this mode, the state space of the
abstract automaton would be infinite. Therefore, to learn this model 
with AIDE, we limit the system to send no more than $N$ packets, i.e.,
we assume that the buffer will not exceed the maximum size of $N$
messages.   
In addition to a standard ``Read'' method which exists in normal ports,
a \textit{non-blocking read} feature is also available
in buffered ports. Identification  results for ports in various
configurations 
are presented in Table~\ref{tab:res1} (top)\footnote{All the 
  experiments in this Section have been carried out on a Sony Vaio
  laptop with 
  Core2Duo 2.26GHz CPU and 4GB of RAM on Ubuntu 12.04.}.  
We have also extended the alphabet of buffered ports to include
non-blocking reading from the port for both strict and non-strict mode 
of reading --- these two experiments are presented with an ``*'' in
Table~\ref{tab:res1}. The reported measures include number of states $|Q|$ and
transitions $|T|$ of the identified model, number of output (``\#MM'') and
equivalence (``\#EQ'') queries in the learning algorithm, number of experiments
on the SUL, and total time spent on learning. 
The behavior of a normal port is similar to a non-strict buffered port.
In normal ports, both reading from port and writing to it 
are blocking, whereas in non-strict buffered port, writing is not a
blocking primitive: if the buffer is not empty, the second write will
overwrite the previous message.  
In buffered ports with strict mode enabled, writing to the port is
non-blocking, but the difference compared to non-strict mode is that
the buffer does not drop older packets and it acts as a
first-in-first-out buffer. In this case, the state space of the model
would be infinite, and thus the buffer should be limited to a maximum
size for finite state identification to work. These behavioral
differences account for the different number of states and transitions
in the identified models, as reported in Table~\ref{tab:res1}.
We have also considered the effects of increasing the maximum 
buffer size ($N$) in buffered ports when the reading mode is set to be
strict. These results are shown in Table~\ref{tab:res1} (bottom). For
  $N=1$, the observed model is the same as normal ports, and by
  increasing the size, the size of the model grows
  gradually. Notice that, if $e(N)$ denotes the number of experiments
  as a function of buffer size $N$, then we see that $e(N)$ grows more
  than proportionally with $n$. 
The CPU time spent for identification grows even faster due to some
overheads in our current implementation. In particular, the capability
of resetting the SUL is required by the identification algorithm. 
In our case, resetting the SUL includes releasing all the resources of the
system, i.e., ports and threads working with them, and initializing
the system again. This operation is performend in the system
wrapper, and, in all our experiments, more than 95\% of the total
identification time is spent to reset the SUL. We expect that 
working on this bottleneck should enable us to experiment with larger
buffer sizes, and also to infer a more accurate growth estimate for
$e(n)$. Another issue which might affect the efficiency is the TCP/IP
connection through the Wrapper Proxy, whereas calling YARP functions
directly in AIDE would slightly decrease the identification time.

\begin{table}[t!]
\begin{center}
\caption{\label{tab:res1}The result of inference for different models
  in YARP port component (top), and inference for different maximum size
of buffer ($N$) in YARP buffered port (bottom). In the topmost table,
for buffered ports, we consider $N=3$.}
\scriptsize
\begin{tabular}{|l|c|c|c|c|c|r|}\hline
\multicolumn{1}{|c|}{Model} & $|Q|$ & $|T|$ & \#MM & \#EQ & \#Experiments & \multicolumn{1}{|c|}{Time (s$\times 1000$)}\\ \hline\hline
port 						& 4 & 5  & 18 & 1 & 33 & 0.6 \\ \hline
buffered port (non-strict) 	& 4 & 6  & 35 & 2 & 50 & 1.0 \\ \hline
buffered port (strict) 		& 8 & 11 & 132 & 4 & 180 & 2.0 \\ \hline
buffered port (non-strict)* & 4 & 8  & 63 & 2 & 114 & 2.5 \\ \hline 
buffered port (strict)*		& 8 & 16 & 224 & 4 & 323 & 12.0
\\ \hline
\multicolumn{7}{c}{\ } \\
\multicolumn{7}{c}{\ } \\
\end{tabular}

\begin{tabular}{|c|c|c|c|c|c|r|}\hline
Size & $|Q|$ & $|T|$ & \#MM & \#EQ & \#Experiments & \multicolumn{1}{|c|}{Time (s$\times 1000$)}\\ \hline\hline
1 & 4  &  5  & 18  & 1 & 32  & 1.0\\ \hline
2 & 6  &  8  & 54  & 3 & 86  & 2.1\\ \hline
3 & 8  &  11 & 132 & 4 & 180 & 2.0 \\ \hline 
4 & 10 &  14 & 225 & 6 & 260 & 10.0 \\ \hline 
5 & 12 &  17 & 504 & 7 & 396 & 17.0 \\ \hline
6 & 14 &  20 & 987 & 8 & 531 & 31.0 \\ \hline 
\end{tabular}
\end{center}
\end{table}

\paragraph{Technical remarks} Since the inferred models are
deterministic, identical queries should produce the same
answer. Therefore, we can cache the queries to avoid expensive
repetitions. This is done by storing a tree of execution traces which
can be exploited to avoid an experiment on a system whenever the
corresponding query is a prefix of another one which has been already
executed. In  our implementation, the cache query is used as a filter
between the MAT and the SUL, which makes it transparent from the
MAT's point of view. Our reports above include only the actual
number of queries on the system. Furthermore, in all of our
experiments, less than 0.5\% of the identification time was spent in
the learning algorithm. The most time-consuming parts are 
network communication, system reset, and thread management. As we have
mentioned above, one reason for such inefficiencies is that the
wrapper uses several delays to make sure it is obtaining the correct
result, since obtaining even one wrong observation in the 
output or equivalence queries would result in failing to learn a
correct model.

\subsection{Verification}
\label{subsec:verification}

The conversion of IAs inferred by AIDE into
a PROMELA model is accomplished as follows. Every model is translated
into one process type which communicates with two unbuffered
channels --- to simulate synchronous communication. These are 
\textit{InChannel} and \textit{OutChannel}, and their task
is to receive input actions from environment, and emit output events
correspondingly. To make the composition flexible, the input and
output channels are the parameters of the process type. At any time,
the next state is determined by the received input action (or the
emitted event) and all the  transitions are performed as atomic
actions. In addition to PROMELA, AIDE is able to export inferred
automata in DOT graph format, C++ and the input language of other
model checkers. The model of programs which use the ports are
translated into automata as well. Here, the translation is manual, but
in principle, it could be done in an automated fashion.  
Finally, the composition of the inferred model with the
code model is done automatically by SPIN.

\paragraph{Case study 1} 
The results of verification, including
the number of generated states by the model checker, the consumed
memory, time and result of verification, are reported 
in Table~\ref{tab:spin1} (top). Considering the 
identified model of a buffered port with non-strict mode of reading,
SPIN finds a deadlock in the model after
exploring 39 states, although the whole state space has about 16K
states. In fact the problem arises if the client sends packets too
quickly through a YARP port configured for non-strict mode. In this
case, there is a concrete chance that the server misses some of the
packets, and a deadlock occurs. For strict mode, we consider a
specific size of buffer $N$, namely $N=1$ and $N=6$.  

\begin{table}[t]
\begin{center}
\caption{\label{tab:spin1}Results of SPIN for the first case study
  (top) and second case  study (bottom). In the topmost table, for
  buffered port, non-strict mode of reading and strict mode of reading
  for ($N=1$ and $N=6$). In the bottommost table, different $N_i$'s and
  size of buffer. The last row is the result of model checking
  with non-blocking read from $Q'_1$} 
\scriptsize
\begin{tabular}{|c|c|c|r|c|}\hline
Model & \#States & Memory(MB) & Time(s) & Conclusion \\ \hline\hline
Buffered Port (non-strict)	& 38		& 128	&	0.01 & deadlock \\ \hline 
Buffered Port (strict, $N=1$)& 15K	& 129	& 0.04	& OK \\ \hline	
Buffered Port (strict, $N=6$)	& 42K	& 132	& 0.09	& OK \\ \hline
\multicolumn{5}{c}{\ } \\
\end{tabular}

\begin{tabular}{|c|c|c|c|c|r|c|c|}\hline
$N_1$ & $N_2$ & $N_3$ & $Size$ & \#States & \multicolumn{1}{|c|}{Time(s)} & Memory(MB) & Conclusion \\ \hline\hline
100   & 100   & 100 & 1 & 8K   & 0.02 & 129 & OK \\ \hline
90    & 100   & 100 & 1 & 790  & 0.01 & 128 & deadlock \\ \hline
100   & 100   & 100 & 6 & 128K & 0.33 & 140 & OK \\ \hline
90    & 100   & 100 & 6 & 1930 & 0.02 & 128 & deadlock \\ \hline
200   & 200   & 200 & 6 & 519K & 3.29 & 176 & OK \\ \hline
180   & 200   & 200 & 6 & 3820 & 0.04 & 129 & deadlock \\ \hline
90$*$ & 100   & 100 & 6 & 19M  & 91.00   & 1300& OK \\ \hline 
\end{tabular}

\end{center}
\end{table}

\paragraph{Case study 2} As before, 
we perform verification for buffered port with strict mode of
reading. The results are presented in Table~\ref{tab:spin1}
(bottom). We examined different values for $N_1$, $N_2$ and $N_3$ and
the size of buffer. As shown in the Table, when $N1=N2=N3$ 
all processes finish successfully. But if either $N1$ or $N2$
are less than $N3$, $P_3$ will be stuck as expected. 
Indeed, in situations where $P_1$ may finish sooner than $P_3$, the
solution would be to change reading from $Q_1'$ (line 5 in Figure
\ref{fig:codeYARP2}) to a non-blocking read. Using the corresponding 
model and the maximum buffer size of 6, SPIN can prove the
non-existence of deadlock in 91 CPU seconds --- last row
of Table~\ref{tab:spin1} (bottom).
%%%%%%%%%%%%%%%%%%%%%%%%%%%%%%%%%%%%%%%%%%%%%%%%%%%%%%%%%%%%%%%%%%
\section{Conclusion}
\label{sec:conc}

In this paper, we show how to exploit automata-based inference and
verification techniques to identify port components of YARP
middleware, and to verify control software build on top of them.
Since YARP is the middleware of choice in the humanoid iCub, 
AIDE can enable the adoption of precise techniques for testing and
verification of relevant components in iCub's control
architecture. 

To the best of our knowledge, this is the first time that a
combination of identification and verification techniques is applied
successfully in robotics. Similar contributions appeared in a series
of works by Doron Peled and others --- see, e.g.,~\cite{adaptivemc06}
for the most recent work --- with hardware verification as the main
target. However, our approach is more general since it decouples
identification techniques from verification techniques, and it enables
the combination of different flavors of such techniques. 

Considering the current limitations of our work, we see
(non)determinism and scalability as the two main issues. As for
nondeterminism, it is well known that middleware can respond in
different ways according to external events which are never completely
under control. The algorithms that we have considered here assume that
the middleware is behaving deterministically, which might turn out to
be an unrealistic assumption. However, in a recent
contribution~\cite{nStart2014}, we have shown how to deal with
nondeterminism when learning Mealy machines, and we expect to be able
to extend this result also to IAs. 
Scaling to more complex components is a challenge for 
our future research agenda. In spite of harsh computational-complexity
results, both identification and verification tools have a record of
success stories in dealing with industrial-sized systems. Furthermore,
AIDE already enables  developers to check their code against common
errors such as, e.g., incorrect port flagging, and it has also been
useful in supplying YARP creators with corner bugs that helped them to
improve some basic functionality of platform. 
We expect that improving the bottlenecks due to resetting the system, 
i.e., managing ports and related threads in the system
wrapper, will improve the capacity of our techniques.  

\bibliographystyle{splncs}
\bibliography{../../AliBib,../../mybib}

\begin{thebibliography}{10}
\providecommand{\url}[1]{\texttt{#1}}
\providecommand{\urlprefix}{URL }

\bibitem{learningIOA2010}
Aarts, F., Vaandrager, F.: Learning {I/O} automata. CONCUR 2010-Concurrency
  Theory pp. 71--85 (2010)

\bibitem{angluin1987}
Angluin, D.: Learning regular sets from queries and counterexamples.
  Information and computation  75(2),  87--106 (1987)

\bibitem{baier2008principles}
Baier, C., Katoen, J.: Principles of model checking. MIT press Cambridge (2008)

\bibitem{cla86a}
Clarke, E., Emerson, E., Sistla, A.: {Automatic verification of finite-state
  concurrent systems using temporal logic specifications}. ACM Transactions on
  Programming Languages and Systems (TOPLAS)  8(2),  263 (1986)

\bibitem{interfaceAutomata2001}
De~Alfaro, L., Henzinger, T.: Interface automata. ACM SIGSOFT Software
  Engineering Notes  26(5),  109--120 (2001)

\bibitem{yarpiCub2008}
Fitzpatrick, P., Metta, G., Natale, L.: Towards long-lived robot genes.
  Robotics and Autonomous systems  56(1),  29--45 (2008)

\bibitem{adaptivemc06}
Groce, A., Peled, D., Yannakakis, M.: Adaptive model checking. Logic Journal of
  IGPL  14(5),  729--744 (2006)

\bibitem{holzmann2004spin}
Holzmann, G.J.: The SPIN model checker: Primer and reference manual, vol. 1003.
  Addison-Wesley Reading (2004)

\bibitem{kearns1994}
Kearns, M., Vazirani, U.: An introduction to computational learning theory. MIT
  press (1994)

\bibitem{nStart2014}
Khalili, A., Tacchella, A.: Learning nondeterministic mealy machines. In:
  Proceedings of the $12^{th}$ International Conference on Grammatical
  Inference (ICGI) (2014), to appear

\bibitem{ioAutomata1987}
Lynch, N.A., Tuttle, M.R.: Hierarchical correctness proofs for distributed
  algorithms. In: Proceedings of the sixth annual ACM Symposium on Principles
  of distributed computing. pp. 137--151. ACM (1987)

\bibitem{metta2010icub}
Metta, G., Natale, L., Nori, F., Sandini, G., Vernon, D., Fadiga, L., von
  Hofsten, C., Rosander, K., Lopes, M., Santos-Victor, J., et~al.: {The iCub
  Humanoid Robot: An Open-Systems Platform for Research in Cognitive
  Development}. Neural networks: the official journal of the International
  Neural Network Society  (2010)

\bibitem{pratt2013darpa}
Pratt, G., Manzo, J.: {The DARPA Robotics Challenge [Competitions]}. Robotics
  \& Automation Magazine, IEEE  20(2),  10--12 (2013)

\bibitem{que82a}
Queille, J., Sifakis, J.: {Specification and verification of concurrent systems
  in CESAR}. In: International Symposium on Programming. pp. 337--351. Springer
  (1982)

\bibitem{quigley2009ros}
Quigley, M., Conley, K., Gerkey, B., Faust, J., Foote, T., Leibs, J., Wheeler,
  R., Ng, A.Y.: {ROS: an open-source Robot Operating System}. In: ICRA workshop
  on open source software. vol.~3 (2009)

\bibitem{shahbaz2008}
Shahbaz, M.: Reverse Engineering Enhanced State Models of Black Box Software
  Components to Support Integration Testing. Ph.D. thesis, Institut
  Polytechnique de Grenoble, Grenoble, France (2008)

\end{thebibliography}
\end{document}